# Towards Secure and Private Language Models for Nuclear Power Plants

**Muhammad Anwar [12], Mishca de Costa [12], Issam Hammad [2] and Daniel Lau [12]**
[1] Data Analytics and AI, Digital Technology and Services, Ontario Power Generation, Pickering, Ontario, Canada
[2] Department of Engineering Mathematics and Internetworking, Faculty of Engineering, Dalhousie University, Halifax, Nova Scotia
muhammad.anwar@opg.com

## Abstract

This paper introduces a domain-specific Large Language Model for nuclear applications, built from the publicly accessible *Essential CANDU* textbook. Drawing on a compact Transformer-based architecture, the model is trained on a single GPU to protect the sensitive data inherent in nuclear operations. Despite relying on a relatively small dataset, it shows encouraging signs of capturing specialized nuclear vocabulary, though the generated text sometimes lacks syntactic coherence. By focusing exclusively on nuclear content, this approach demonstrates the feasibility of in-house LLM solutions that align with rigorous cybersecurity and data confidentiality standards. Early successes in text generation underscore the model's utility for specialized tasks, while also revealing the need for richer corpora, more sophisticated preprocessing, and instruction fine-tuning to enhance domain accuracy. Future directions include extending the dataset to cover diverse nuclear subtopics, refining tokenization to reduce noise, and systematically evaluating the model's readiness for real-world applications in nuclear domain.

## 1. Introduction

The field of Natural Language Processing (NLP) has witnessed significant advances in recent years, largely driven by the emergence of Transformer-based Large Language Models (LLMs) such as the GPT and LLaMA families of models. These models leverage self-attention mechanisms and large datasets to learn complex linguistic patterns, enabling them to generate coherent text, summarize documents, and assist in code generation. However, most state-of-the-art LLMs are trained on large quantities of general-purpose text, often sourced from the internet, making them adept at a wide range of generic tasks but less specialized for niche, high-stakes applications.

In the nuclear industry, domain specificity and data confidentiality are of utmost importance. Nuclear operations, safety protocols, and regulatory documents often contain sensitive information that organizations cannot risk sharing with external parties. Moreover, public cloud infrastructures which are essential for large-scale AI training, pose notable cybersecurity concerns, particularly in a sector where the potential ramifications of data leakage are serious. Current commercial LLM APIs generally operate as black-box services; users submit text queries and receive generated responses without visibility or full control over how their data is processed or stored. Such an approach sometime is not feasible with the strict security requirements upheld in nuclear facilities and research institutions.

To address these challenges, this paper presents a method for training a specialized LLM from scratch on publicly available nuclear domain data, using a single GPU in a secure, on-premises environment. Rather than adapting a pre-trained general-purpose model, a model architecture from scratch is developed inspired by GPT and LLaMA, keeping in mind the specific requirements from the nuclear industry. This approach ensures that the entire training pipeline—from data acquisition and tokenization to model optimization and text generation—remains under the full control of the nuclear organization. By circumventing reliance on external platforms, the risk of cyber threats, inadvertent data leaks, or unauthorized access to confidential information could be considerably reduced.

A key motivation for this work is to demonstrate the feasibility and accessibility of training such a model in a resource-constrained setting, specifically on a single GPU workstation. Many nuclear organizations do not have access to large-scale cloud clusters or specialized data centers. Thus, the focus of this chapter is to explore large language model training that is very accessible to other nuclear organizations due to low cost and resource requirement. While this paper mainly focuses on nuclear applications, the framework can be generalized to other high-security, domain-specific sectors such as defense, finance, or healthcare.

Figure 1 outlines the steps to train a large language model, which will be explained in detail in this paper.

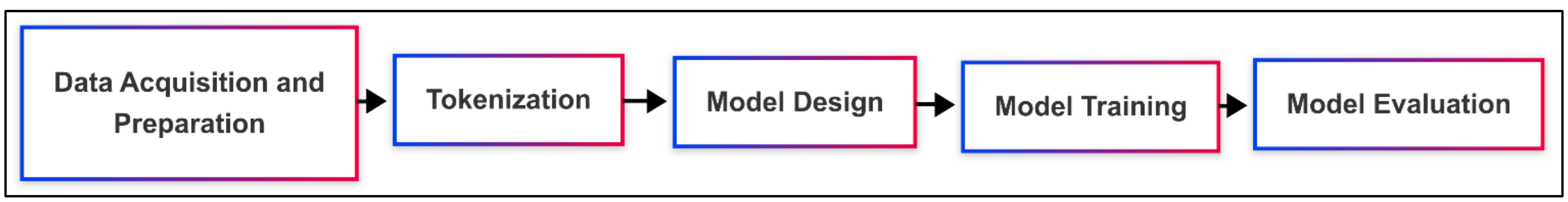

**Figure 1 Training steps**

## 2. Related Work and Literature Review

Modern Large Language Models (LLMs) owe much of their progress to the Transformer architecture, introduced by Vaswani et al. in their groundbreaking work “Attention Is All You Need” [1 This architecture moved away from the Recurrent Neural Network (RNN) and Convolutional Neural Network (CNN) architectures, in favor of self-attention mechanisms, which enable efficient learning of long-range dependencies within text. Subsequent research, including GPT-3 by Brown et al. [2], built on the transformer architecture to demonstrate the remarkable capabilities of large-scale language modeling through training on large volumes of data. GPT-3’s success across a wide variety of NLP tasks, often with minimal or no task-specific fine-tuning, underscored the fundamental strength and versatility of the transformer design.

With the Transformer serving as a solid foundation, researchers have increasingly focused on pre-training LLMs on large text corpora before adapting them to specific tasks. Early demonstrations of this approach, such as GPT-2 [3], emphasized the effectiveness of large, unsupervised pre-training in capturing detailed language representations. The subsequent development of large datasets like The Pile, an 800GB collection of heterogeneous text [4], further highlights the importance of diverse training corpora for fostering broad language understanding and robust generalization. These large-scale efforts have established a paradigm wherein models benefit from exposure to extensive and varied text, laying the groundwork for enhancing downstream performance. In short, the key to training highly capable LLMs is to feed them large volumes of diverse data.

Despite the success of general-purpose LLMs, resource efficiency remains a significant concern, especially for organizations with limited computational infrastructure or strict data security requirements. Methods like GaLore [5] aim to address the high memory and computational costs associated with training large models. By optimizing memory usage and computational steps, such techniques can make advanced LLMs more accessible, even within specialized or regulated domains where large-scale cloud computing may be infeasible.

The benefits of Transformer-based models are also evident in domain-specific applications. Research exploring large-model capabilities in high-stakes fields, like healthcare, demonstrates that with the right fine-tuning, LLMs can achieve high-level performance in tasks such as medical question answering [6]. Similarly, BloombergGPT [7] shows how specialized data, and fine-tuning can yield a finance-focused LLM adept at tasks ranging from market analysis to financial report summarization. These advancements suggest that, given appropriate training data and security provisions, LLMs can serve as valuable tools across a range of specialized industries—including the nuclear sector, which stands to benefit from the controlled, on-premises training approaches outlined in my work.

## 3. Data Acquisition and Preparation

This section details the important first phase for developing the specialized nuclear domain LLM: data acquisition and preparation. Raw textual information, especially from specialized fields like nuclear, cannot be directly ingested by these models. Instead, it requires a journey from its original state to a processed format ready for model training.

### 3.1 Data Sources

A important first step in building a domain-specific Large Language Model (LLM) for the nuclear sector is assembling a good-quality corpus that accurately reflects both the technical language and serves as a source of knowledge of nuclear domain. In this study, the primary data source was the *Essential CANDU* textbook [8], an open-access resource provided by the University Network of Excellence in Nuclear Engineering (UNENE). This textbook offers comprehensive coverage of the CANDU (CANada Deuterium Uranium) reactor technology, from foundational nuclear physics to reactor operation and safety practices.

The *Essential CANDU* textbook spans over 20 chapters covering topics such as:

- Historical Context and Reactor Evolution: Chapters on CANDU's genealogy and development
- Reactor Physics and Dynamics: In-depth discussions of neutron physics, reactor statics, and reactor kinetics
- Thermal-Hydraulics and Plant Systems: Design principles for heat transport, steam generation, and plant operations
- Instrumentation, Control, and Electrical Systems: Architectural design of control systems and major electrical components of nuclear plants
- Radiation Protection, Safety, and Regulation: Comprehensive treatments of radiation monitoring, regulatory frameworks, and safety analysis methods

- Nuclear Plant Materials, Chemistry, and Fuel: Materials selection, corrosion issues, reactor chemistry, and various fuel cycles
- Waste Management and Fuel Handling: Storage, disposal of irradiated fuel, and on-power refueling details
- In-Core Fuel Management: Mathematical models and computational methods for managing reactor fuel

Because each chapter focuses on different facets of CANDU reactor technology, the textbook provides a valuable and varied corpus—ranging from theoretical principles to practical design, operational strategies, and regulatory requirements. This breadth is intended to enable the LLM to learn domain-specific terminology (e.g., "pressure tubes," "delayed neutrons," "moderator system") and contextual nuances important for comprehension within the Canadian nuclear landscape.

### 3.2 Preprocessing and Cleaning

Given that the "Essential CANDU" textbook is well-structured and relatively free of any irrelevant text, a minimal preprocessing approach is adopted to preserve the original language and context. The main preprocessing steps are detailed in the table below:

**Table 1: Pre-processing Steps**

| Preprocessing Step | Description |
|---|---|
| **Text Extraction** | Each chapter's PDF is converted into raw text. Diagrams, tables, and embedded images were omitted or replaced with short placeholder text with the relevant titles. |
| **Header and Footer Removal** | Standard document headers, footers, and page numbers are stripped out to avoid unnecessary repetition or noise. |
| **Basic Line Merging** | Merged lines and paragraphs where needed to ensure a more natural textual flow suitable for sentence-based tokenization. |
| **Retention of Chapter Structure** | Apart from removing non-informative elements (like navigation text), the textbook's headings and organizational markers were left intact, providing contextual cues that might assist the model in understanding topical transitions. |

Because each chapter of "Essential CANDU" is authored or co-authored by recognized nuclear subject matter experts, a minimal approach to data cleaning is utilized to keep most of the original content unchanged.

### 3.3 Tokenization and Embedding Strategy

A key challenge in enabling computers to process and "understand" human language is converting text, which is inherently symbolic and qualitative, into a numerical format upon which mathematical models can operate. This conversion process involves two primary stages: tokenization and embedding.

To illustrate, consider teaching a computer to read some specialized technical content, such as the "Essential CANDU" textbook. The initial step is to segment the continuous stream of text into manageable pieces, like how humans perceive sentences as composed of words. This segmentation is called 'tokenization'. An advanced tokenizer, however, might break text into slightly different units called tokens. These tokens could be whole words (e.g., "reactor," "neutron"), common sub-word units (e.g., "ion" in "ionization," or "-ing" in "processing"), or even individual characters if a word is very rare or unfamiliar. The objective is to create a vocabulary of these tokens that can efficiently represent the entire corpus. For instance, the term "CANDU" might become a single token, while a very long or rare technical term might be represented as a sequence of a few sub-word tokens. This approach ensures that common, important terms are treated as whole units, while the system can still handle new or rare words without needing an excessively large dictionary. Once the text is segmented into these tokens, each unique token in the vocabulary is assigned a unique numerical identifier, a token ID (e.g., "CANDU" might be ID 500, "reactor" ID 501, the sub-word "moder" ID 502, and "ator" ID 503, so "moderator" becomes the sequence [502, 503]). This step transforms the entire textbook into a long sequence of these integer IDs.

While token IDs provide a numerical representation, these integers don't inherently capture any meaning or relationship between tokens. For example, the ID 500 for "CANDU" and 501 for "reactor" doesn't tell the model that these terms are semantically related in the nuclear domain. This is where embeddings come in. An embedding is a way to represent each token ID as a list of numbers, called a vector, in a multi-dimensional space. The core idea is that these vectors are not arbitrary; the model learns to assign vector values such that tokens with similar meanings or that are used in similar contexts in the nuclear textbook will have vectors that are "close" to each other in this multi-dimensional space (e.g., the vector for "neutron" might be mathematically similar to the vector for "proton" but different from the vector for "turbine"). This learned numerical representation allows the model to perform mathematical computations that reflect semantic relationships.

Furthermore, since the meaning of text heavily depends on word order, which standard token embeddings do not inherently capture, positional embeddings are incorporated. These are additional vectors that provide the model with information about the absolute or relative position of each token within the input sequence

### 3.4 Dataset Structure

Once the data processing steps described in previous sections is completed, a single unified corpus from all chapters of the "Essential CANDU" textbook is compiled. This corpus is then divided into three subsets to facilitate model development and evaluation:

- Training Set (≈ 80%): This portion contained most of the text, spanning various chapters, and was used for the primary training of the model.
- Validation Set (≈ 10%): This subset was used periodically during the training process to fine-tune hyperparameters and to detect potential overfitting of the model to the training data.
- Test Set (≈ 10%): This subset was held out exclusively for the final performance evaluation of the model. Its purpose is to ensure that the model's ability to generalize is assessed on unseen text covering topics such as reactor physics, operations, safety, and regulatory matters.

### 3.5 Ethical and Legal Considerations

The use of the "Essential CANDU" textbook, which is published online under open-access or permissible licensing, ensures that the risks of proprietary data breaches or regulatory non-compliance related to this data source are low. This approach also aligns with the needs of nuclear organizations seeking to replicate or adapt this methodology while upholding stringent data governance. No sensitive or export-controlled data was incorporated into this dataset, reinforcing the reproducibility of the pipeline and maintaining the model's focus on openly available nuclear knowledge.

With the data acquisition and preparation stages for the "Essential CANDU" textbook corpus now detailed, the next section will describe the transformer-based architecture and training methodology that was employed to develop a domain-specific LLM suited to the demands of the nuclear industry.

## 4. Model Architecture

### 4.1 Overview and Influences

The design of this implementation of Large Language Model (LLM) follows the transformer architecture, a commonly used model architecture that powers well-known models like GPT and LLaMA. The architecture implementation of the LLM draws from Sebastian Raschka's Build a Large Language Model (from Scratch) [9]. A key aspect of this implementation is that this model is being constructed from its fundamental components, rather than adapting a large, pre-trained general-purpose model. This from-scratch approach allows us to tune the transformer's essential building blocks specifically for the nuclear domain, using the Essential CANDU textbook as our core dataset.

**Table 2: Model Parameters**

| **Parameter** | **Value** | **Description** |
|---|---|---|
| Vocabulary Size | 50257 | Sets the total number of unique tokens that the model can recognize, covering both domain-specific and general text. |
| Context Length | 256 | Determines the maximum number of tokens processed in a single forward pass, defining how much text can be handled at once. |
| Embedding Dimension | 768 | Specifies the dimensionality of the token embeddings, influencing the representational capacity of each token vector. |
| Number of Attention Heads | 12 | Indicates how many parallel attention mechanisms the model uses to analyze different aspects of the input sequence. |
| Number of Transformer Layers | 12 | Establishes how many stacked Transformer blocks the model will have, each block containing attention and feed-forward layers. |
| Dropout Rate | 0.1 | Represents the probability of randomly dropping connections during training, which helps prevent overfitting. |
| Query Key Value (QKV) Bias | False | Governs whether a bias term is added to the queries, keys, and values in the attention mechanism. |

## 4.2 Core Building Blocks

### 4.2.1 Vector Embeddings (Token & Positional)

Vector embeddings serve as the foundation for the model's understanding of textual elements. Each word or sub-word in the text is represented as a high-dimensional numerical vector, often referred to as an embedding. This approach captures certain facets of meaning and usage, allowing the model to learn specialized vocabulary such as moderator, PHWR, or CANDU 6 directly from the nuclear-domain corpus. Because transformers do not inherently account for word order, positional embeddings are also introduced to encode the position of each token within a sequence. This numeric signal enables the model to distinguish between tokens at the beginning of a sentence and those appearing later, ensuring that it can track references from one paragraph to the next.

### 4.2.2 Multi-Head Self-Attention

Once the combined token and positional embeddings have been computed for each token in an input sequence, the 'self-attention mechanism' is the core component that allows the model to understand the relationships between different tokens within that same sequence. The intuition is that the meaning of a word is often determined by its context. For example, in the phrase "the primary heat transport system uses heavy water as a coolant," to understand the role of "heavy water," the model needs to "attend" to words like "coolant" and "heat transport system." Self-attention enables each token to look at all other tokens in the input sequence (up to the context length) and assign scores indicating how relevant each other token is for interpreting the current one.

The process described above constitutes a single "attention head." Multi-head self-attention, as used in our model (12 heads), involves performing this self-attention mechanism multiple times in parallel. Each "head" can potentially learn to focus on different types of relationships or different aspects of the input sequence (e.g., one head might focus on syntactic dependencies, another on semantic similarity related to nuclear components). The outputs from these multiple heads are then typically concatenated together and passed through another linear transformation to produce the final output of the multi-head attention block. This allows the model to capture a richer set of relationships simultaneously, which is crucial in a specialized nuclear context for linking reactor design parameters to their units, associating operational constraints with safety guidelines, or consistently recalling acronyms and references introduced in earlier chapters.

### 4.2.3 Feed-Forward Layers and Residual Connections

After attention has been computed, each token's representation is fed into a position-wise feed-forward network. This two-layer neural network refines the attended information and transforms it into a more abstract representation. Residual connections, sometimes referred to as skip-connections, enable the model to preserve core information from earlier layers by passing the input directly to the output of each block. This design helps stabilize training, as it prevents essential context from being lost. Layer normalization then maintains outputs within a stable numerical range, improving the reliability of training and helping the model converge more efficiently.

### 4.3 Architectural Configuration

In practice, transformer-based models build upon these components—attention, feed-forward networks, residual connections, and normalization—by stacking them into multiple layers or blocks. The depth and width of these layers have a direct impact on both model capacity and the computational resources required. In this paper, a moderate number of layers was selected to balance the expressive power needed for nuclear-domain language against the limitations of training on a single GPU. Multiple parallel attention heads in each layer capture different perspectives of token relationships, and the dimensionality of token embeddings is chosen to be large enough for intricate domain terms yet small enough to fit into GPU memory.

### 4.4 Hyperparameters: The "Dials" of Model Training

Hyperparameters in machine learning represent preset configurations that influence model behavior without being directly learned during training. In this setup, they act like tuning knobs that reconcile computational constraints with the model's complexity and performance goals. One central hyperparameter is the learning rate, which dictates the size of parameter updates after each training step. A higher learning rate accelerates convergence but can sometimes cause oscillations or instability, whereas a lower rate may yield more consistent improvements at the cost of increased training time. Another key hyperparameter is the batch size, referring to how many data samples the model processes before updating its parameters. While larger batches often produce smoother gradient estimates, smaller batches mitigate the high memory demands characteristic of training on a single GPU. Lastly, dropout probability determines how frequently parts of the model's internal connections are temporarily inactivated, a measure designed to prevent overfitting on domain-specific data that may not be extensive in size.

### 4.5 Domain-Specific Considerations

Unlike broad internet-scale language models, this work targets a narrower nuclear corpus, which presents distinct benefits and challenges for the model's architecture and training. The token distribution becomes more specialized, with certain technical terms such as "moderator," "fusion," or "burnup" appearing with notably high frequencies. This density of domain-specific expressions can reshape how the embedding vectors evolve during training, potentially leading to more refined representations for these domain specific terms.

In addition, the effective context length (256 in our setup) is a key parameter. Nuclear literature often involves protracted discussions, detailed procedural steps, or multi-step derivations that require understanding relationships over extended passages of text. While a larger context length would be ideal for capturing these very long-range dependencies, the choice of 256 tokens reflects a necessary trade-off for a model of this size to remain trainable on a single GPU. This constraint directly impacts the maximum number of prior tokens the self-attention mechanism can consider when computing the context for a given token.

## 5. Training Methodology

Developing a specialized Large Language Model (LLM) for nuclear domain requires a careful balance between the technical demands of language modeling and the resource constraints imposed by a single-GPU setup. In the following subsections, we detail our end-to-end training strategy, from defining the model's predictive objective to the specifics of optimization and performance monitoring.

### 5.1 Next-Token Prediction as the Objective

During training, next-token prediction is the central training objective for our LLM. The model learns the patterns of the nuclear language by repeatedly trying to predict the next token (word or sub-word) in a sequence, given all the tokens that came before it. In other words, when the model is presented with a sequence of tokens from the *Essential CANDU* textbook, for each token in that sequence, it attempts to infer the most probable token that should follow.

### 5.2 Training Process

Each complete pass over the entire training dataset is termed an epoch. During an epoch, the text data is divided into smaller mini-batches—sequences of tokens of a manageable size for our single-GPU setup (e.g., a batch might contain a few sequences, each 256 tokens long). For each mini-batch, the model performs a forward pass: it takes the input tokens and generates predictions (probability distributions over the vocabulary) for the next token at every position in those sequences.

The difference between these model predictions and the actual true tokens from the textbook is then quantified using a loss function. For next-token prediction tasks, the standard loss function is cross-entropy loss. This function measures how "different" or incorrect the model's predictions are. If the model assigns a high probability to the correct next token, the cross-entropy loss will be low; if it assigns a low probability, the loss will be high. The goal of training is to minimize this loss.

After calculating the loss for a mini-batch, the backpropagation algorithm is employed. This algorithm efficiently computes the gradients of the loss function with respect to all the model's learnable parameters (the weights and biases in its layers). These gradients essentially indicate the direction and magnitude by which each parameter should be adjusted to reduce the loss. An optimizer then uses these gradients to update the model parameters. This cycle of forward pass, loss calculation, backpropagation, and parameter update is repeated for all mini-batches in the training set, and then this entire process is repeated for multiple epochs, allowing the model to iteratively refine its parameters and improve its predictive accuracy.

### 5.3 Performance Monitoring and Preventing Overfitting

Throughout the training process, it's crucial to monitor not only the model's loss on the training set but also its performance on the separate validation set. A common metric for this, closely related to cross-entropy loss, is called perplexity. Perplexity provides a more intuitive measure of how well the language model predicts the sample of text from the validation set. A lower perplexity score indicates that the model is less "surprised" or uncertain about the sequence of tokens in the validation data, meaning its predictions are more accurate. It is directly derived from the cross-entropy loss.

By tracking both training and validation loss (or perplexity), we can detect overfitting. Overfitting occurs when the model starts to perform very well on the training data (memorizing it, including its noise) but its performance on the unseen validation data stagnates or even degrades. This is a sign that the model is not learning generalizable patterns. When overfitting is observed—typically seen as a diverging trend where training loss continues to decrease while validation loss starts to increase—we may need to adjust regularization strategies (like dropout strength) or consider early stopping, where training is halted once performance on the validation set no longer improves significantly.

Figure 2 shows the loss curves which shows the drop in training and validation loss throughout the training progression.

### 5.4 Single-GPU Constraints and Security Measures

A notable feature of this method is the compact model design, specifically sized so that both training and inference can be performed on a single GPU. This capability is especially important in nuclear settings, where reliance on large-scale cloud infrastructure may be undesirable for certain applications due to stringent security regulations. By ensuring the model remains small enough to fit on commonly available GPU hardware, this approach makes it practical for nuclear facilities to train and serve the model entirely in-house.

Moreover, this self-contained approach is replicable in an offline or air-gapped environment, allowing sensitive nuclear data to remain private throughout the development pipeline. Because the training loop does not rely on external servers or cloud-based components, organizations can maintain more stringent access controls, thereby preventing data leakage or unauthorized access

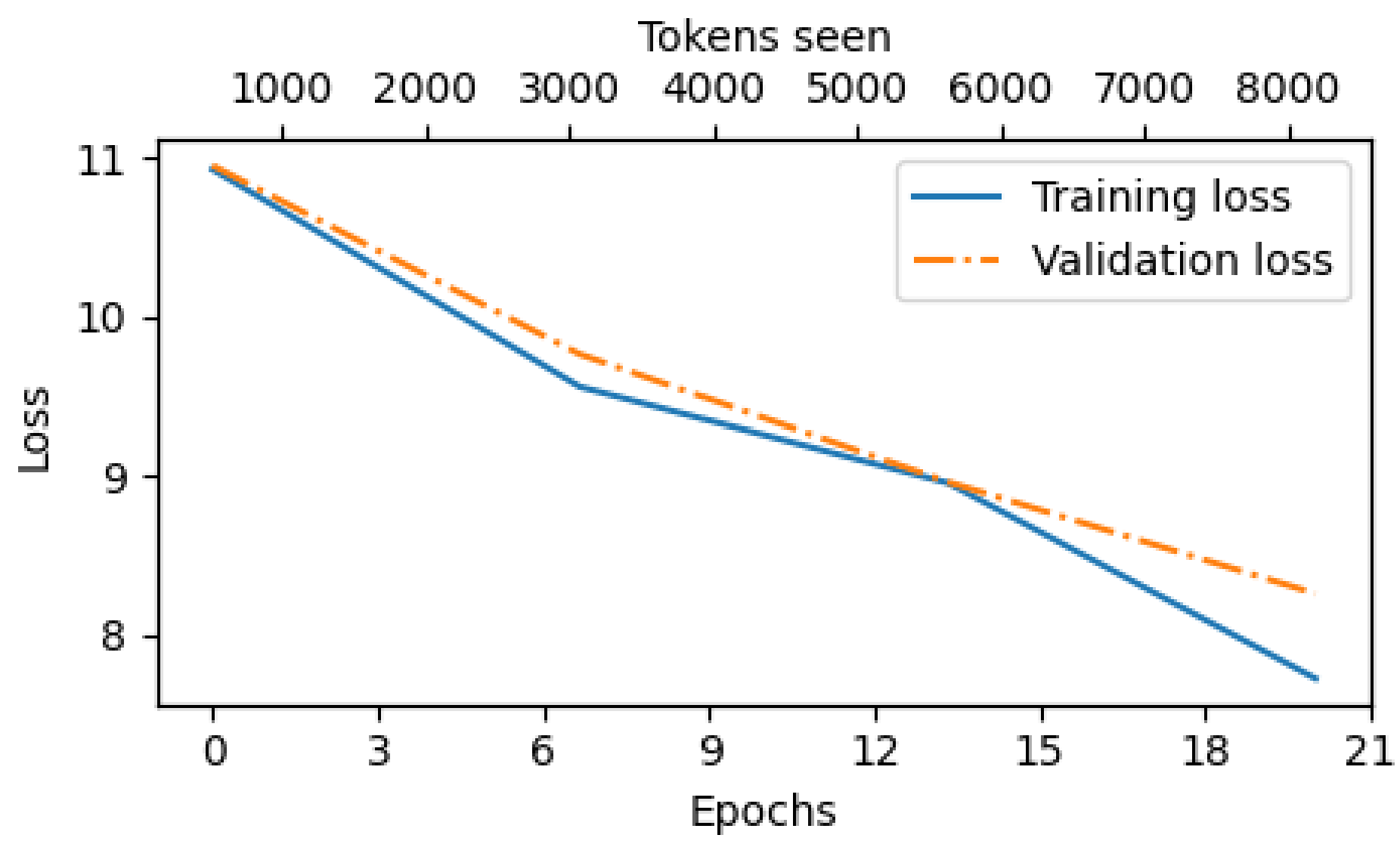

**Figure 2: Training and Validation loss curves**

## 6. Evaluation and Results

To gauge our model's performance after 20 epochs of training on the *Essential CANDU* textbook, we conducted a simple test by prompting the model with a series of start contexts related to nuclear-domain topics. Each prompt represents an incomplete sentence—such as "Different types of commercial reactors like BWRs and" or "distinctive features of CANDU reactors such as horizontal fuel"—and asks the model to generate a continuation. Although the resultant texts are far from fluent or contextually complete, the model's outputs do exhibit recognizable nuclear terminology and concepts. For example, words like "Darlington," "decay reactions," and "nuclear force" occasionally appear in relevant (though fragmented) contexts.

A closer look at these generated segments reveals that, while the model's training is insufficient to produce coherent extended text, it has indeed assimilated some domain-specific vocabulary. Phrases referring to reactor types, neutron interactions, and fundamental nuclear forces confirm that the network has begun to associate certain keywords with relevant technical discussions. Despite this progress, the overall prose remains inconsistent and often loses logical coherence after a few tokens.

Such limitations are expected, given the compact nature of the model and the relatively small, noisy dataset. Training was conducted under a modest experimental setup on a laptop, with minimal preprocessing and a single data source. The fact that the model can nonetheless reproduce some nuclear-sector lexicon is a promising indicator for further development. With more extensive data and longer training regimes, one can anticipate more contextually aligned and fluent outputs, potentially making the model a valuable resource for summarizing technical documents or assisting in specialized nuclear research.

## 7. Conclusion and Future work

This project represents an initial step toward creating a domain-specific Large Language Model aimed at nuclear applications. Although the results show limited coherence in generated text, the model clearly demonstrates partial mastery of specialized nuclear phrases and concepts. This early traction underscores the suitability of the chosen architecture and methodology for nuclear-focused LLM development, even under modest computational resources.

Central to these findings is the importance of data quality. Moving forward, improved text extraction pipelines, possibly leveraging advanced optical character recognition (OCR) tools or domain-aware parsers, would help provide cleaner and more structurally consistent input. By minimizing noise and irrelevant text, subsequent training iterations should yield more logically coherent generations.

Another strategic avenue for enhancing the model involves continued pretraining. One approach would be to take a small, pretrained model—like GPT-2—and further train it on a curated corpus of nuclear-domain texts. Because such a model already possesses a foundation in general language understanding, it would likely converge faster and with greater accuracy on nuclear-specific knowledge. This strategy could be augmented by instruction fine-tuning, where the model learns to respond more directly to human-posed queries or to follow technical instructions consistently. These steps have the potential to transform the LLM into a more robust, adaptive tool for generating summaries, clarifying complex procedures, or supporting decision-making in nuclear settings.

In conclusion, the work presented here lays the groundwork for a specialized LLM that can eventually find applications in document summarization, operational planning, and knowledge management for the nuclear sector. The next phases—ranging from more advanced data curation to extended pretraining—offer a clear path to a model that not only understands the specialized terminology but also communicates it fluently and contextually. By continuing to refine and expand this framework, researchers and practitioners can unlock the potential of LLMs as secure, on-premises assets in the highly regulated world of nuclear science and engineering.

## 8. Acknowledgments

This research paper was supported by Ontario Power Generation (OPG) and by The Natural Sciences and Engineering Research Council of Canada (NSERC) and The Canadian Nuclear Safety Commission (CNSC) grant number ALLRP 580442-2022.